# World Conference on Scientific Discovery and Innovation 2023,
*May 24–26, 2023, Florida, USA*

# MACHINE LEARNING FOR FRAUD DETECTION IN DIGITAL BANKING: A SYSTEMATIC LITERATURE REVIEW

**Md Zahin Hossain George[1]; Md Khorshed Alam[2]; Md Tarek Hasan[3];**

[1]. BSc in Csc, Daffodil International University, Dhaka , Bangladesh;
Email: mdzahinhossaingeorge@gmail.com
[2]. BSc in Csc, Daffodil International University, Dhaka , Bangladesh;
Email: mdkhorshed950101@gmail.com
[3]. BSc in Electrical and Electronics Engineering, Daffodil International University, Dhaka , Bangladesh;
Email: mdtarekhasan58749@gmail.com

**Abstract**
This systematic literature review examines the role of machine learning in fraud detection within digital banking, synthesizing evidence from 118 peer-reviewed studies and institutional reports. Following the PRISMA guidelines, the review applied a structured identification, screening, eligibility, and inclusion process to ensure methodological rigor and transparency. The findings reveal that supervised learning methods, such as decision trees, logistic regression, and support vector machines, remain the dominant paradigm due to their interpretability and established performance, while unsupervised anomaly detection approaches are increasingly adopted to address novel fraud patterns in highly imbalanced datasets. Deep learning architectures, particularly recurrent and convolutional neural networks, have emerged as transformative tools capable of modeling sequential transaction data and detecting complex fraud typologies, though challenges of interpretability and real-time deployment persist. Hybrid models that combine supervised, unsupervised, and deep learning strategies demonstrate superior adaptability and detection accuracy, highlighting their potential as convergent solutions. The review further underscores the importance of evaluation metrics—precision, recall, F1-score, and PR-AUC—as well as explainability frameworks like SHAP and LIME, which ensure that models are both statistically robust and operationally transparent. Cross-regional analysis shows that regulatory environments and institutional capacities shape methodological adoption: the European Union emphasizes compliance under PSD2 and GDPR, North America leverages fintech partnerships and data-driven innovation, and emerging economies rely heavily on infrastructure and governance maturity. Despite methodological advances, gaps remain in reproducibility, robustness under distributional shift, and theoretical integration with criminological and governance frameworks.

## INTRODUCTION

Fraud in digital banking can be broadly defined as intentional deception perpetrated through electronic financial systems with the goal of illicit financial gain or unauthorized access to resources. According to Hashemi et al. (2022), fraud is traditionally conceptualized as a deviation from expected behavior that undermines the integrity of financial transactions. In the context of digital banking, fraud extends beyond classical forms of misrepresentation to encompass activities such as phishing, account takeover, synthetic identity creation, and transaction laundering. Researchers such as Thar and Wai (2024) emphasize that digital fraud is distinct in its reliance on rapid, high-volume, and often cross-border electronic transactions, making traditional detection strategies insufficient. Legal and institutional definitions also frame fraud as a systemic risk: Nair et al. (2025) describes it as a threat to the safety and soundness of banking systems, Priya and Saradha (2021) documents its role in undermining customer confidence in digital payment infrastructures. Fraud's conceptual boundaries, therefore, include not only theft of assets but also manipulation of digital identities and exploitation of institutional vulnerabilities. The academic literature converges on the idea that defining fraud in digital banking requires accounting for both behavioral irregularities and the technological mediums through which these crimes occur (Achary & Shelke, 2023).

**Figure 1: Digital Banking Fraud Detection Framework**

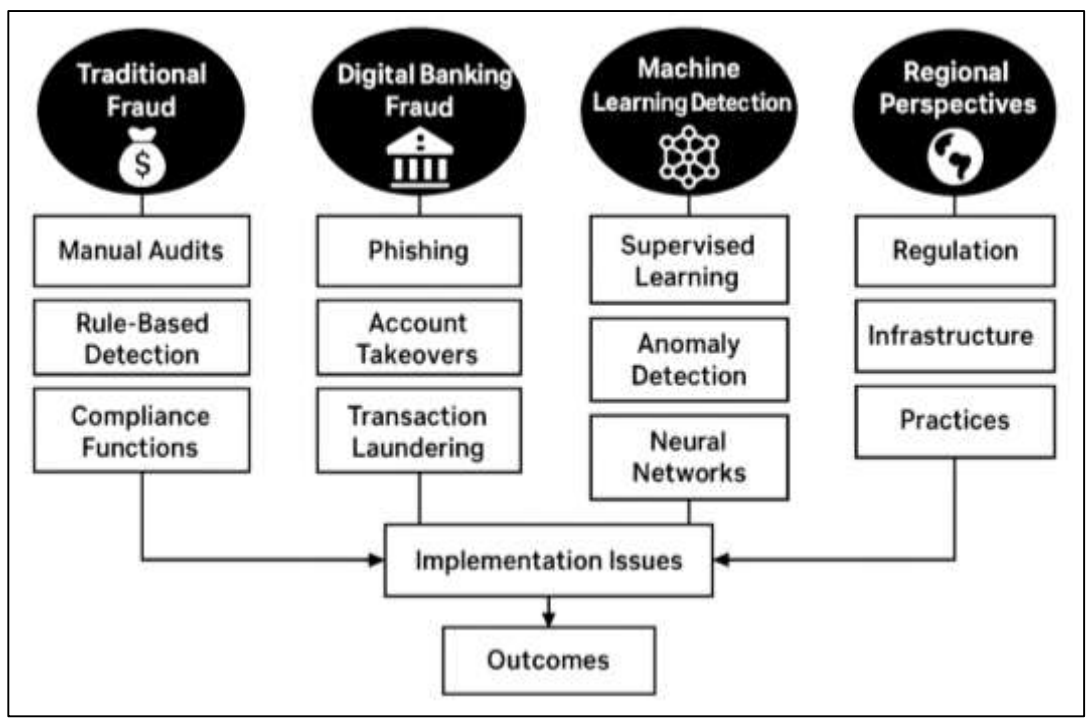

The global significance of fraud detection in digital banking is underscored by its direct economic impact, erosion of trust, and implications for financial stability. The International Monetary Fund reports that cyber-enabled fraud has become one of the most pressing risks facing financial institutions worldwide, threatening not only individual banks but also the stability of interconnected financial systems. Similarly, the World Bank identifies digital fraud as a barrier to financial inclusion in developing economies, where rapid adoption of mobile and online banking platforms often outpaces the development of fraud-prevention infrastructure. Empirical studies highlight the magnitude of losses: Swathi et al. (2024) note that billions of dollars are lost annually through credit card and online banking fraud, the systemic cost implications of fraudulent activity in large-scale financial ecosystems. Fraud reduces consumer willingness to adopt digital financial products, hindering broader goals of digital transformation. Furthermore, cross-border fraudulent schemes exploit gaps in jurisdictional authority and regulatory enforcement. The literature confirms that fraud detection in digital banking is not confined to organizational profitability but is integral to the maintenance of global economic resilience and financial stability.

**Figure 2: Process Flow of Fraud Detection Systems**

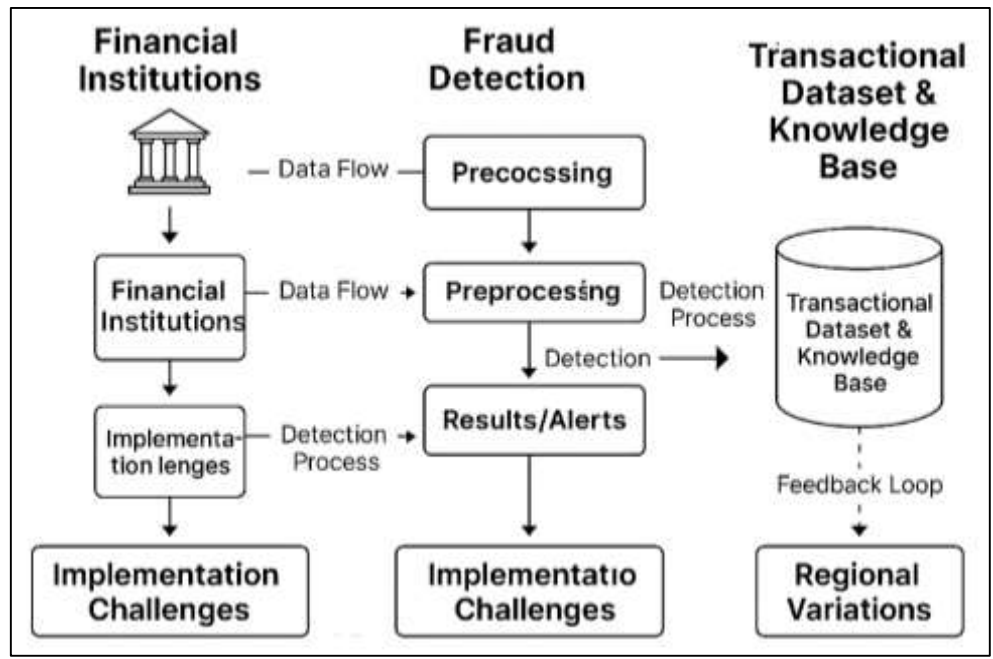

Historically, fraud detection in financial systems relied heavily on manual audits, expert judgment, and rule-based mechanisms. Early systems used predefined thresholds, blacklists, and spending limits to detect unusual transactions. While such rule-based approaches were transparent and interpretable, they suffered from rigidity and high false-positive rates, particularly as transaction volumes increased. Chhabra and Prabhakaran (2023a) illustrate that rule-based mechanisms were unable to adapt dynamically to evolving fraud tactics, necessitating more flexible data-driven solutions. Rule-based systems were insufficient to keep pace with fraudsters' ability to manipulate thresholds and exploit static detection rules. This historical evolution underscores the limitations of traditional methods in addressing modern fraud, particularly in digital banking contexts where real-time transaction monitoring is essential. The recognition of these limitations provided the impetus for the adoption of advanced statistical models and, eventually, machine learning approaches that can learn from data and adapt to changing fraud patterns.

Machine learning emerged as a powerful alternative to rule-based systems by offering the ability to learn patterns, identify anomalies, and generalize across evolving fraud tactics. Ngai et al. (2011) identify supervised learning methods—such as decision trees, logistic regression, and support vector machines—as early applications in fraud detection, demonstrating superior performance in classification tasks. Susto et al. (2018) highlight that anomaly detection techniques extended detection capabilities to contexts with scarce or imbalanced labeled data. The role of ensemble models and deep learning architectures in capturing nonlinear and sequential fraud behaviors. Neural networks, in particular, have been applied to sequential transaction streams, with recurrent and convolutional models yielding strong performance in identifying suspicious activity. Scholars such as Al-dahasi et al. (2025) stress that machine learning represents a paradigm shift by moving detection from static compliance functions toward adaptive, data-driven processes. The literature confirms that machine learning is now a cornerstone of fraud detection in digital banking, offering scalability and predictive accuracy that surpass earlier methods.

Although machine learning has enhanced fraud detection capabilities, technical and operational challenges remain central to the literature. A primary concern is the imbalance of fraud datasets, where fraudulent transactions represent only a small fraction of total data, complicating the ability of classifiers to generalize (Li & Chen, 2025). Techniques such as oversampling, cost-sensitive learning, and anomaly detection have been developed to address this challenge. Privacy and data protection regulations also restrict the sharing of sensitive transaction data for model training, creating limitations in dataset diversity and quality. False positives are another operational challenge: Yekollu et al. (2024)

report that excessive false alarms increase operational costs and erode customer trust. Adversarial manipulation, where fraudsters deliberately craft inputs to evade detection, further complicates implementation. Scholars agree that these challenges highlight the complexity of applying machine learning in high-stakes, real-time financial systems.

Comparative analyses reveal that fraud detection approaches differ across regions due to variations in regulation, infrastructure, and institutional practices. In the European Union, PSD2 and strong customer authentication requirements drive the use of risk-based models that comply with regulatory thresholds. In North America, innovation is driven by fintech partnerships and large-scale data collaborations, with institutions emphasizing scalability and cloud-based fraud detection systems. Emerging economies, in contrast, face challenges related to limited infrastructure, insufficient regulatory enforcement, and high reliance on mobile banking platforms. Studies by Kalkan et al. (2020) emphasize that infrastructural and cultural differences influence the effectiveness of fraud detection technologies. The literature demonstrates that fraud detection cannot be viewed solely as a technical issue but must be contextualized within regional financial ecosystems. The breadth and diversity of fraud detection research highlight the necessity of systematic synthesis to consolidate fragmented findings. Previous reviews, such as those by Elavarasan et al. (2020), provided valuable overviews but lacked the scope to integrate emerging advances in deep learning, hybrid models, and interpretability frameworks. More recent syntheses, such as Fainshmidt et al. (2020), have focused on specific methodological aspects but do not provide comprehensive cross-regional or cross-methodological comparisons. argue that without systematic synthesis, research remains fragmented, hindering the development of coherent frameworks for application. The inclusion of diverse studies in this review spanning supervised, unsupervised, deep learning, hybrid models, and regulatory perspectives ensures that a wide range of methodological and contextual insights are integrated. This approach enables a more holistic understanding of how machine learning contributes to fraud detection in digital banking and highlights the academic value of structured evidence mapping (Lee & Wright, 2016).

## LITERATURE REVIEW

The literature on fraud detection in digital banking reveals a multidisciplinary convergence of financial risk management, computer science, and regulatory policy. Scholars have consistently demonstrated that fraud detection cannot be examined in isolation; rather, it must be understood within the broader framework of digital transformation in banking systems. Early studies primarily relied on statistical and rule-based approaches to identify fraudulent patterns, but the exponential growth of transaction volumes and sophistication of fraud tactics necessitated the adoption of machine learning techniques. Recent contributions show that machine learning not only improves classification accuracy but also enables real-time, large-scale fraud detection across international markets. Furthermore, literature emphasizes that the efficacy of fraud detection systems depends not only on algorithmic innovations but also on data quality, organizational culture, and regulatory compliance. This section of the review synthesizes scholarship from multiple disciplines to provide an integrated understanding of machine learning applications in fraud detection. The subsections that follow are organized thematically, beginning with the conceptual and historical underpinnings of fraud detection research, progressing through algorithmic techniques, implementation challenges, and comparative cross-regional insights. By structuring the literature along these lines, the review establishes conceptual clarity and facilitates the systematic mapping of knowledge in the field.

### Fraud in Financial Services

Defining fraud in the context of financial services requires consideration of both legal interpretations and technological dimensions. Scholars consistently describe fraud as intentional deception designed to secure unlawful gains, often involving misrepresentation or concealment of information. In digital banking, fraud encompasses a spectrum of illicit behaviors, including account takeover, synthetic identity creation, phishing, and transaction manipulation, all of which exploit vulnerabilities in online financial systems. From a conceptual perspective, fraud boundaries extend beyond theft to include practices such as money laundering and terrorist financing, which undermine institutional and regulatory integrity. The difficulty of establishing clear definitions lies in the evolving nature of fraud schemes and the overlap between financial crime, cybercrime, and insider abuse. Nizioł (2021) frame fraud within broader systemic risk frameworks, underscoring its potential to destabilize public trust and financial markets. Academically, studies emphasize that conceptual clarity in defining fraud is essential to designing effective detection models, as vague or inconsistent definitions hinder comparability across research and practice. Thus, the literature reflects ongoing efforts to delimit fraud conceptually while acknowledging its multifaceted and adaptive nature.

**Figure 3: Fraud Detection Evolution in Finance**

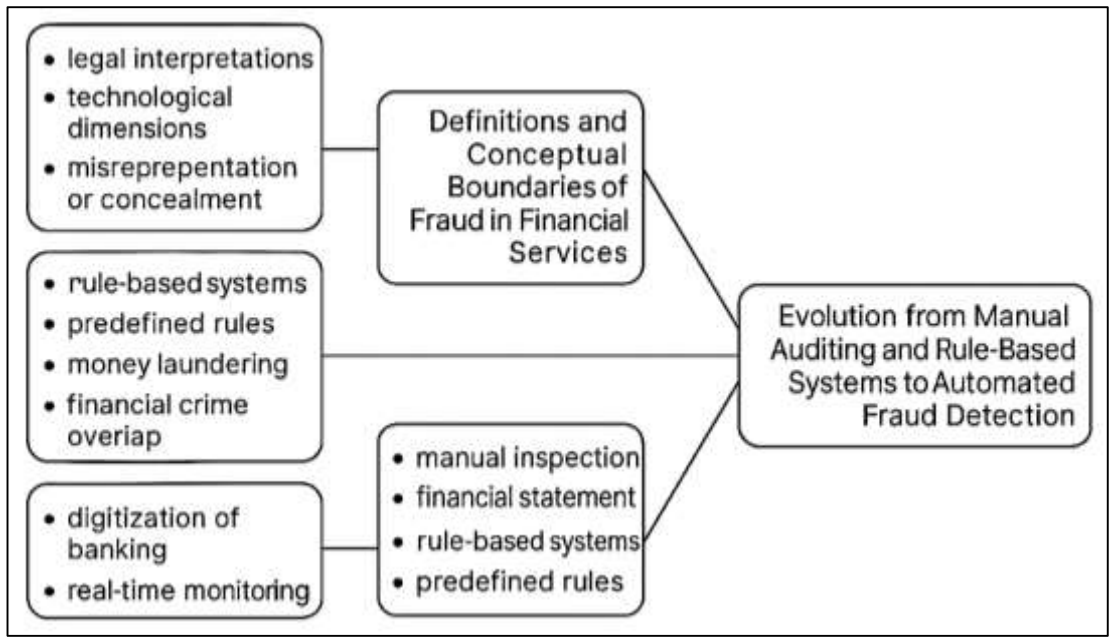

Fraud detection in financial services has undergone a significant evolution from traditional manual auditing practices to automated, technology-driven systems designed to address the scale and complexity of modern banking. In the earliest phase, fraud detection relied heavily on manual inspection of financial statements, reconciliations, and compliance checks, where auditors identified anomalies through accounting expertise and transaction-level scrutiny. While these methods were valuable for small-scale operations, they became inadequate as transaction volumes increased exponentially in the digital era. The introduction of rule-based systems in the 1980s and 1990s marked an early attempt at automation, using predefined rules such as transaction thresholds, spending limits, or blacklisted accounts to flag suspicious activities (Jahid, 2022; Ridzuan et al., 2024). Despite their interpretability, rule-based approaches were criticized for rigidity, inability to adapt to new fraud patterns, and high false-positive rates (Arifur & Noor, 2022). The transition to automated fraud detection was accelerated by the digitization of banking services, which enabled real-time transaction monitoring and introduced statistical and machine learning techniques into fraud detection research. Studies highlight that automated approaches offered the scalability and adaptability required to handle the diversity of fraud tactics in online banking, marking a paradigm shift in fraud detection from compliance-centered practices to data-driven analytics (Hasan & Uddin, 2022).

Fraud detection in digital banking operates at the intersection of technological, organizational, and legal boundaries, each shaping the way fraudulent activities are understood and addressed. From a technological standpoint, fraud detection encompasses algorithms, data processing pipelines, and transaction monitoring infrastructures designed to identify anomalies (Rahaman, 2022). Organizationally, fraud detection is embedded in risk management practices, compliance mandates, and customer service operations that dictate how alerts are interpreted and acted upon. Legal frameworks introduce another layer, as fraud definitions and enforcement mechanisms vary across jurisdictions, influencing the scope of detection systems and their compliance obligations. The literature stresses that fraud detection boundaries must consider the challenges of cross-border financial flows, regulatory heterogeneity, and data privacy constraints, all of which complicate the design of universally applicable solutions (Halbouni et al., 2016; Md Rahaman & Ashraf, 2022).

Moreover, researchers highlight that fraud detection should not be conflated with fraud prevention; detection refers specifically to the identification and classification of fraudulent transactions after they occur, whereas prevention encompasses broader deterrence and security strategies. This distinction, while subtle, is critical in setting conceptual boundaries for academic research and institutional practice.

**Machine Learning Paradigms in Fraud Detection**

Supervised learning methods form the foundation of machine learning applications in fraud detection, as they leverage labeled datasets where fraudulent and legitimate transactions are known. Decision trees have long been used due to their interpretability and ability to handle heterogeneous variables, with Quinlan's (1993) C4.5 algorithm being among the earliest applied to banking data (Islam, 2022). Logistic regression, although simple, remains popular for its statistical grounding and clear interpretability in predicting binary fraud outcomes. Support vector machines (SVMs) are widely recognized for their robustness in high-dimensional spaces, providing accurate separation between fraudulent and legitimate transactions, particularly when kernel methods are employed (Hasan et al., 2022). Ensemble models such as random forests and gradient boosting machines improve classification accuracy by reducing variance and bias, often outperforming single classifiers in empirical studies. Deep learning techniques, including multilayer perceptrons, have also been tested for supervised fraud detection tasks, with Vynokurova et al. (2020) reporting strong performance in real-world transaction datasets. Studies consistently highlight that supervised learning models depend heavily on high-quality labeled data, yet they offer superior accuracy when sufficient training samples are available. Overall, supervised classification-based methods remain dominant in financial fraud detection research due to their proven effectiveness in predictive modeling (Redwanul & Zafor, 2022).

**Figure 4: Fraud Detection Machine Learning Framework**

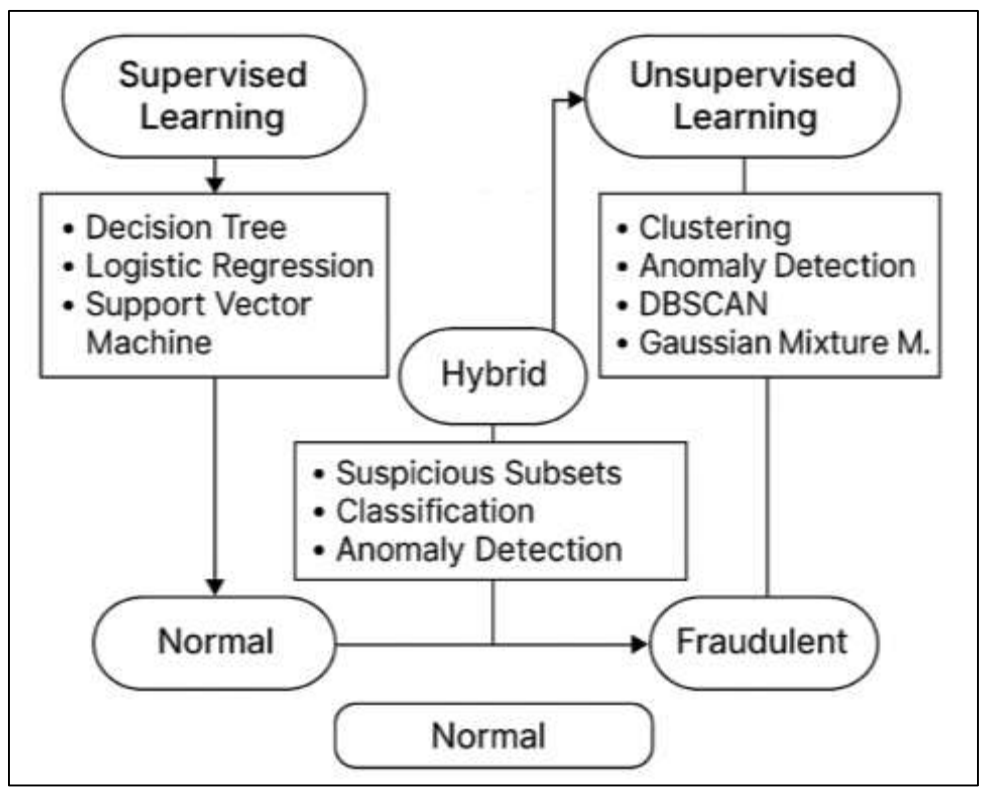

Unsupervised learning approaches, particularly anomaly detection and clustering, are indispensable for fraud detection in contexts where labeled datasets are scarce or outdated. Fraudulent transactions often represent a small fraction of the data, making supervised training challenging; anomaly detection conceptualizes fraud as statistical deviations from normal patterns (Rezaul & Mesbaul, 2022). Clustering methods such as k-means, hierarchical clustering, and self-organizing maps group similar transactions together, enabling identification of outliers that may signal fraud (Hasan, 2022). Density-based methods such as DBSCAN have also been effective in highlighting suspicious clusters of unusual transactions. Probabilistic models, including Gaussian mixture models, provide another layer of anomaly detection by estimating the likelihood of a transaction under a distribution of legitimate behaviors. In banking environments where fraud tactics evolve rapidly, unsupervised learning provides the flexibility to capture previously unseen fraudulent behaviors (Tarek, 2022). However, studies also emphasize the limitations of unsupervised methods, particularly high false-positive rates, which can burden banking operations with unnecessary alerts. Despite these challenges, the literature consistently supports anomaly detection as a critical paradigm for adaptive fraud detection in digital banking.

Hybrid frameworks that integrate supervised and unsupervised methods have emerged as a promising paradigm for addressing the shortcomings of each approach in isolation. Research indicates that supervised models perform well when sufficient labeled data exists, while unsupervised models are advantageous in detecting novel fraud types. Hybrid systems combine the strengths of both paradigms by first using unsupervised clustering or anomaly detection to identify suspicious subsets of data, which are then classified by supervised models (Kamrul & Omar, 2022). Carcillo et al. (2021) demonstrated that hybrid approaches reduce false positives while improving recall, thereby enhancing operational efficiency. Ensemble-based hybrids, such as stacking or cascading supervised and unsupervised algorithms, further optimize classification accuracy and adaptability. In addition, semi-supervised frameworks, which train models on a small labeled dataset alongside large volumes of unlabeled data, have proven particularly effective in fraud detection where labeled fraud cases are scarce. Hybrid systems have been implemented in real-time monitoring environments, where anomaly detection narrows the search space and supervised classifiers provide precision (Kamrul & Tarek, 2022). Literature across financial data mining emphasizes that hybrid approaches offer a balanced solution to fraud detection's dual challenge: accurately detecting known fraud while adapting to emerging schemes (Mubashir & Abdul, 2022).

Comparative analyses across supervised, unsupervised, and hybrid paradigms underscore the trade-offs inherent in machine learning approaches to fraud detection. Supervised learning models, such as logistic regression, decision trees, and SVMs, consistently achieve high accuracy when trained on balanced, high-quality labeled data (Muhammad & Kamrul, 2022). However, their reliance on labeled examples limits adaptability in fast-evolving fraud landscapes. Unsupervised learning, by contrast, excels in discovering novel fraud behaviors and handling unlabeled datasets but is often hampered by high false-positive rates and interpretability issues (Reduanul & Shoeb, 2022). Hybrid models mitigate these issues by integrating strengths of both paradigms, with empirical evidence showing improved precision, recall, and operational feasibility. Studies such as those by Malik et al. (2022) emphasize that performance differences depend significantly on data availability, fraud prevalence, and the specific banking context. Furthermore, comparative literature highlights the growing emphasis on interpretability and explainability as critical factors, particularly when ML systems are embedded into regulated financial institutions (Kumar & Zobayer, 2022). Thus, the body of research presents fraud detection paradigms not as competing methodologies but as complementary strategies that collectively enhance the resilience of digital banking systems.

**Deep Learning Architectures and Neural Networks in Transaction Monitoring**

Deep learning architectures have become central to transaction monitoring because they model temporal dependence and local regularities that characterize card swipes, money transfers, and mobile payments. Recurrent neural networks (RNNs) and their gated variants—Long Short-Term Memory (LSTM) and Gated Recurrent Units (GRU)—encode the chronology of user behavior, enabling the model to condition each decision on a history of prior events (Sadia & Shaiful, 2022). In payment streams, where fraudulent bursts often follow benign activity, LSTM/GRU layers capture recency effects and seasonality. Convolutional neural networks (CNNs) contribute complementary capacity (Noor & Momena, 2022): 1-D convolutions detect short-range motifs such as rapid merchant switches, unusual time-of-day patterns, or spending spikes, while dilated/stacked filters provide multi-scale receptive fields for heterogeneous cadence. Empirical studies on real card datasets report that RNNs and CNNs can be trained on raw or lightly engineered features, reducing dependence on handcrafted rules while improving stability under shifting transaction mixes (Istiaque et al., 2023). Attention mechanisms layered on top of RNN/CNN backbones further emphasize salient subsequences—e.g., a suspicious merchant-MCC chain—thereby sharpening decision boundaries without exhaustive feature crafting. Across benchmarks, these architectures integrate naturally with cost-sensitive losses and class-imbalance strategies common in fraud analytics. The result is a family of models that learns sequential signatures of misuse while preserving operational throughput required for near-real-time authorization (AlSagri, 2025).

**Figure 5: Deep Learning Fraud Detection Framework**

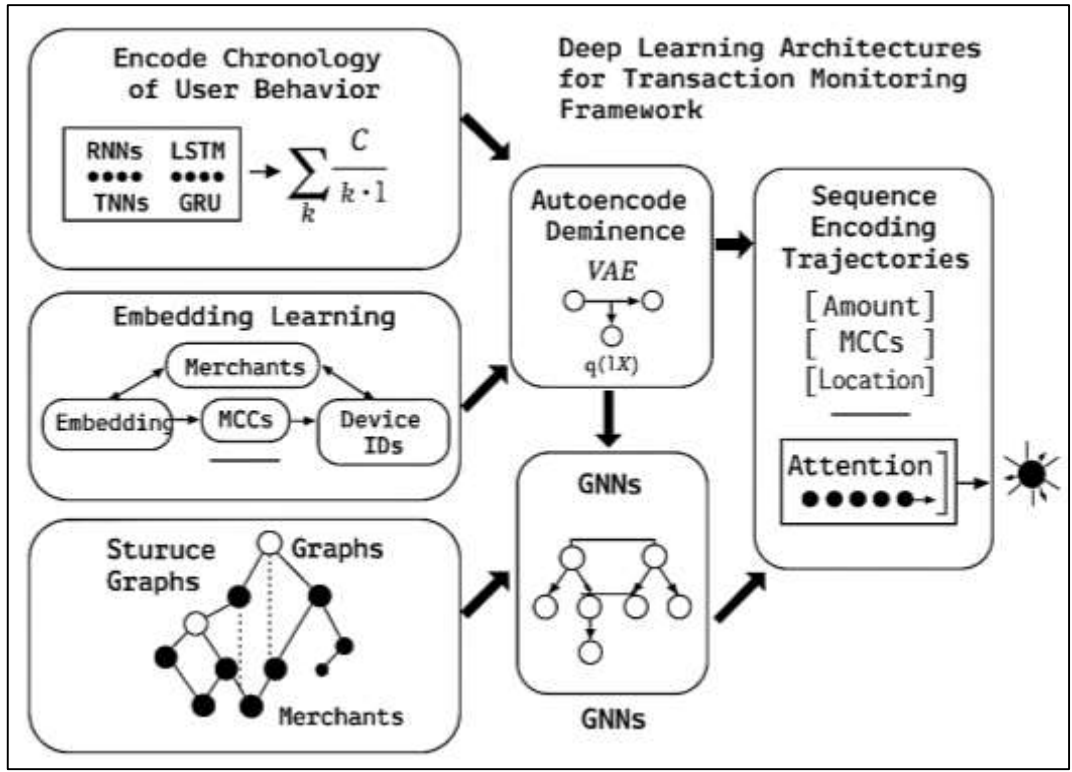

A defining contribution of deep networks to fraud analytics is representation learning—the automatic discovery of feature spaces that separate normal and fraudulent behavior (Hasan et al., 2023). In sequential banking data, embeddings can be learned for merchants, merchant category codes (MCCs), device IDs, and geo-tokens, enabling the system to infer proximity among entities that co-occur in legitimate routines or in coordinated abuse. Autoencoders compress transaction windows into low-dimensional codes where reconstruction error signals deviance; variational autoencoders (VAEs) extend this idea by modeling latent distributions, which supports thresholding by probabilistic rarity. On graph-structured views of banking ecosystems—cardholder–merchant bipartite graphs, device–account linkages, or money-flow networks—graph neural networks (GNNs) propagate signals over neighborhoods to surface organized rings and mule clusters that are not obvious at the individual-transaction level (Hossain et al., 2023). Sequence encoders combine time gaps, amounts, MCCs, and locations to learn trajectory-aware embeddings sensitive to tempo and periodicity. Attention weights and saliency maps provide localized explanations of which events, merchants, or intervals were informative, supporting analyst review and model governance (Sultan et al., 2023). Surveys and domain studies underscore that these learned representations reduce reliance on brittle expert rules and ease adaptation across portfolios with different spending cultures, while maintaining screening accuracy at low fraud base rates.

**Data Imbalance and Real-Time Processing Challenges**

A central methodological challenge in digital banking fraud detection lies in the class imbalance problem, where fraudulent transactions represent only a minute fraction of the total transaction volume. Research consistently demonstrates that conventional classifiers trained on such skewed data tend to favor the majority class, resulting in high overall accuracy but poor recall of fraudulent cases. Oversampling techniques, including random oversampling and the Synthetic Minority Oversampling Technique (SMOTE), have been widely adopted to rebalance datasets by generating synthetic minority class samples. SMOTE and its extensions, such as Borderline-SMOTE and Adaptive Synthetic Sampling (ADASYN), improve minority representation while reducing overfitting compared to naïve duplication (Hossen et al., 2023; Razzaq & Shah, 2025). Conversely, undersampling methods selectively remove majority class samples to achieve class balance, though often at the cost of discarding useful information. Cost-sensitive learning approaches provide an alternative by assigning higher misclassification penalties to fraudulent cases, thereby aligning model optimization with real-world financial risks (Tawfiqul, 2023). Ensemble methods such as EasyEnsemble and BalanceCascade combine undersampling with boosting to enhance detection capability without compromising the integrity of majority class data (Sanjai et al., 2023). Empirical evidence across fraud detection studies highlights that hybrid approaches—combining SMOTE with cost-sensitive ensembles—achieve superior recall while maintaining precision in highly imbalanced digital banking datasets. Collectively, the literature underscores that imbalanced learning is not merely a preprocessing issue but a core methodological priority for fraud detection in financial services (Akter et al., 2023).

**Figure 6: Data Imbalance and Real Time Processing Challenges**

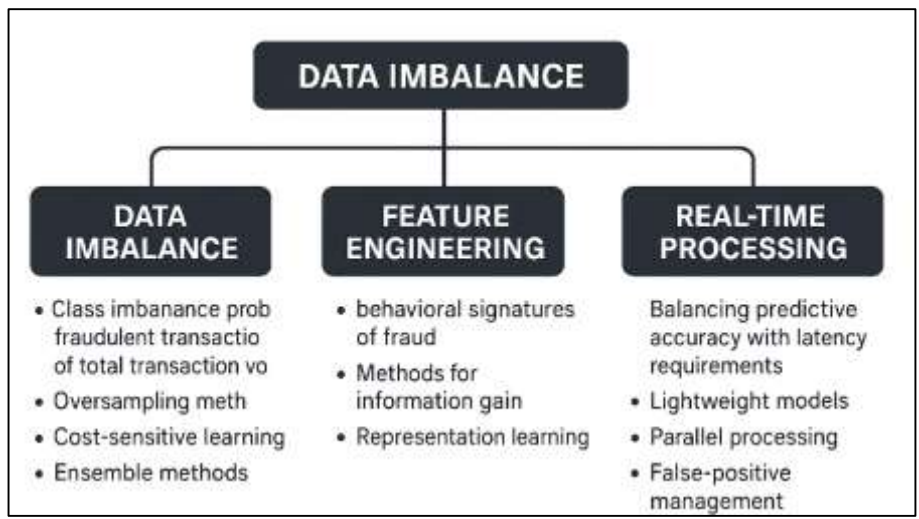

Fraud detection models depend heavily on the quality and relevance of features extracted from raw transaction data. High-dimensional datasets typical in digital banking include transaction amounts, merchant identifiers, geolocation, temporal information, device fingerprints, and customer demographics, often generating hundreds or thousands of variables. Effective feature engineering seeks to capture behavioral signatures of fraud, such as velocity features (number of transactions within a time window), burst spending patterns, or sudden geographic jumps in purchase location. Feature selection methods, including recursive feature elimination, mutual information, and principal component analysis (PCA), reduce dimensionality while preserving predictive power. Wrapper-based methods evaluate subsets of features through iterative model training, whereas embedded techniques such as LASSO regularization incorporate feature selection directly into the learning algorithm. Representation learning through deep autoencoders and graph embeddings provides a data-driven alternative by constructing compact latent spaces without extensive manual engineering. Studies by (Zhang et al., 2025) demonstrate that careful feature engineering enhances classifier robustness, particularly under covariate shifts in customer behavior. Moreover, comparative evaluations show that models with feature-rich transaction contexts consistently outperform those relying on raw attributes alone. The literature highlights that feature engineering is not a one-time step but an iterative process aligned with both fraud patterns and system-specific constraints.

The operational deployment of fraud detection models requires balancing predictive accuracy with stringent real-time constraints imposed by digital banking infrastructures. Large-scale systems such as card authorization networks demand responses within milliseconds, leaving little room for computationally intensive algorithms. Research emphasizes that high-latency models, even if accurate, risk delaying transaction approvals and negatively impacting customer experience. Consequently, fraud detection systems often employ lightweight feature extraction pipelines and optimized models to meet latency requirements. Parallel processing frameworks and distributed computing, such as Apache Spark and GPU acceleration, have been integrated to scale detection capacity across millions of transactions per second. Another major constraint lies in false-positive management: overly sensitive models may flag legitimate transactions, creating unnecessary declines that erode customer trust and revenue. Studies highlight that cost-sensitive thresholding and adaptive feedback loops are critical to balancing fraud detection sensitivity with operational efficiency. Furthermore, compliance requirements such as GDPR and PSD2 introduce restrictions on data storage, processing, and cross-border transfer, adding complexity to real-time fraud detection pipelines (Altalhan et al., 2025). Collectively, the literature indicates that real-time fraud detection must address not only algorithmic efficiency but also regulatory alignment, system scalability, and customer experience.

**Interpretability in Fraud Detection Systems**

Evaluating fraud detection systems requires performance metrics that account for the highly imbalanced nature of financial transaction datasets, where fraudulent events constitute less than 1% of all transactions. Standard metrics such as accuracy can be misleading, as models predicting all transactions as legitimate would achieve near-perfect accuracy while failing to detect actual fraud. Precision and recall have therefore become central indicators: precision measures the proportion of flagged frauds that are truly fraudulent, while recall quantifies the proportion of actual frauds correctly identified. The F1-score, the harmonic mean of precision and recall, balances these dimensions and provides a more holistic evaluation metric.

**Figure 7: Real-Time Banking Fraud Detection Framework**

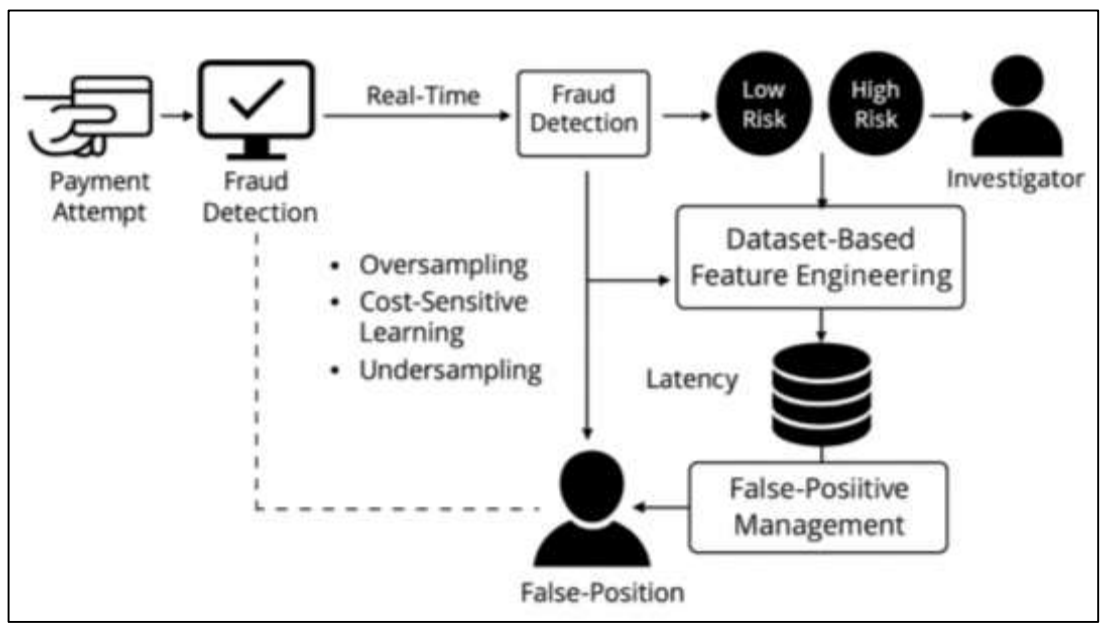

In addition, the area under the receiver operating characteristic curve (ROC-AUC) has been widely used to evaluate classifier discrimination across thresholds, but scholars note that ROC-AUC can overstate performance in imbalanced contexts (Krawczyk, 2016). As a result, the precision-recall area under the curve (PR-AUC) is increasingly preferred for fraud detection tasks, as it emphasizes the minority class performance. Cost-sensitive evaluation frameworks extend these metrics by weighting false negatives more heavily than false positives, aligning model evaluation with financial risk realities. Thus, the literature underscores that the choice of performance metrics is not only statistical but also context-specific, with profound implications for the operational utility of fraud detection systems in digital banking (Kota et al., 2004).

A persistent theme in fraud detection research is the trade-off between maximizing predictive accuracy and ensuring interpretability of machine learning models (Parsa et al., 2019). Traditional models such as logistic regression and decision trees provide transparent decision-making processes, enabling auditors and regulators to understand why a transaction was classified as fraudulent. However, while interpretable, these models often underperform compared to complex ensembles or deep neural networks, which achieve higher detection accuracy but operate as "black boxes". Financial institutions face the challenge of balancing these two dimensions: on the one hand, maximizing fraud capture reduces financial losses, but on the other, lack of transparency complicates regulatory compliance and undermines user trust. Research suggests that interpretability cannot be sacrificed entirely for accuracy, as explainability is essential for mitigating risks of bias, error propagation, and legal liability. Hybrid approaches such as using interpretable surrogate models to approximate black-box outputs and integrating feature importance rankings within ensemble frameworks (Lundberg & Lee, 2017) have been developed to bridge this gap. The literature converges on the idea that fraud detection requires not merely high-performing models but also explainable mechanisms to support decision-making in compliance-driven environments (Awosika et al., 2024).

**Cross-Regional Insights**

European fraud analytics operates within a dense regulatory architecture that directly shapes model design, data access, and operational thresholds. The Revised Payment Services Directive (PSD2) requires open banking via access-to-account interfaces and mandates Strong Customer Authentication (SCA), with the EBA's Regulatory Technical Standards (RTS) specifying multi-factor controls and conditions for Transaction Risk Analysis (TRA) exemptions (Kaslow & Biernaux, 2015). TRA ties authentication leniency to empirically demonstrated fraud rates, which in practice couples machine-learning performance metrics to supervisory tolerances. GDPR further constrains feature construction, lawful bases for processing, and model explainability obligations, thereby pushing institutions toward privacy-preserving pipelines and auditable detection logic. Cross-border SEPA rails and ISO 20022 migrations standardize message fields that feed fraud features, aiding portability of engineered variables across member states. Empirical and review studies note that EU card-not-present fraud remains a key pressure point, encouraging sequential and anomaly-based monitors aligned to SCA step-up policies (Ranta et al., 2018). Post-implementation assessments associate SCA with reductions in unauthorized transactions, while also documenting shifts in fraudster adaptation patterns that sustain the need for TRA-driven recalibration. Operationally, European banks emphasize interpretable models and traceable data lineage to satisfy audit trails, with XAI overlays increasingly used to evidence proportionality and fairness. Academic syntheses underline that EU institutions optimize under multi-objective constraints—fraud loss, user friction, and regulatory compliance—where risk-sensitive thresholds and periodic model validation are embedded in governance. Collectively, PSD2/SCA, GDPR, and pan-European payment harmonization form a regime in which fraud detection is inseparable from supervisory metrics and explainability requirements (Yang et al., 2025).

North American fraud detection reflects a more market-led configuration that prioritizes data-driven experimentation, cloud-native pipelines, and bilateral fintech-bank partnerships. Rather than a single open-banking mandate, U.S. data sharing has evolved through private API ecosystems and industry frameworks, enabling rapid prototyping of graph, sequence, and streaming models on large-scale authorization data. Supervisory expectations focus on model risk governance and safety-and-soundness—e.g., SR 11-7/OCC guidance on validation, challenger testing, and monitoring—which conditions deployment with documentation, back-testing, and concept-drift surveillance.

AML/CFT rules (FinCEN) and sectoral privacy regimes intersect with PCI DSS and NIST digital identity guidance, shaping feature eligibility and authentication orchestration without prescribing a single SCA-style template. Empirical studies describe banks and card networks co-developing ensemble and deep-sequence models, leveraging consortium signals and merchant intelligence to lift precision at fixed review budgets. Fintech partnerships accelerate adoption of graph-based entity resolution, device intelligence, and behavior biometrics in risk scoring, with regulators emphasizing explainability and adverse-action logic when outcomes affect consumers. Comparative work notes higher tolerance for model complexity provided validation evidence is robust and performance gains are material, reinforcing a culture of iterative tuning and A/B experimentation across channels. The literature thus characterizes North America by flexible data collaboration, rigorous model governance, and strong private-sector roles in scaling ML fraud detection (Bartkowski et al., 2023).

**Figure 8: Global Fraud Detection Regulatory Framework**

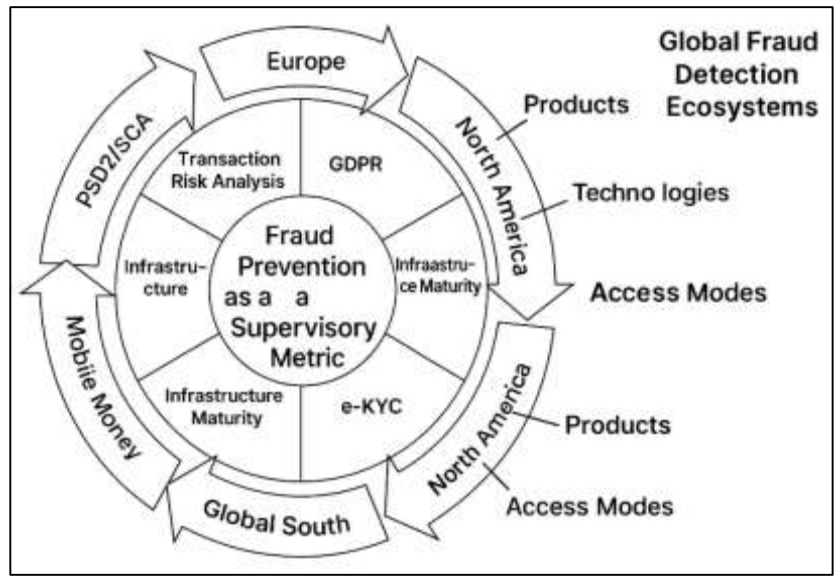

Fraud detection in emerging economies is shaped by mobile-first adoption, cash-digital interlinkages, and heterogeneous institutional capacity. Financial inclusion has expanded through mobile money and fast-payment platforms, generating distinctive transaction topologies—agent networks, peer-to-peer rails, and micro-merchant ecosystems—that alter fraud typologies and feature design. Governance frameworks vary widely: some jurisdictions deploy real-time retail payments with central bank stewardship (e.g., Brazil's Pix; India's UPI), which yields standardized data fields supportive of ML features and centralized fraud intelligence; others rely on fragmented private rails with uneven reporting. Studies document that identity infrastructure—SIM registration, national ID, e-KYC— conditions the feasibility of linking accounts, devices, and beneficiaries, affecting the power of graph-based detection. AML/CFT compliance under FATF standards influences data retention and cross-border information sharing, with mutual evaluations highlighting needs for improved suspicious transaction reporting and analytics capacity. Case literature on South Asia and Sub-Saharan Africa shows that agent collusion, social-engineering rings, and mule networks require hybrid anomaly-and-rule monitors tuned to local usage cycles and liquidity patterns. Research on Bangladesh, Kenya, Mexico, and Nigeria associates improvements in rails and supervision with measurable gains in alert quality and decline management when standardized reference data becomes available. Across studies, infrastructure maturity, data standardization, and supervisory capacity are repeatedly identified as levers determining which ML paradigms—tabular ensembles, sequential encoders, or graph networks—yield operational lift in emerging markets (Laurentis & Pearson, 2021).

**Fraud Detection Models into Banking Ecosystems**

Deployment of fraud detection models in banking ecosystems proceeds through staged pipelines that align data engineering, model risk governance, and production monitoring. Studies describe reference patterns that begin with offline back-testing against historical authorizations, proceed to shadow-mode runs in parallel with incumbent rules, and culminate in canary or phased rollouts keyed to loss exposure and customer segments. Financial institutions formalize this lifecycle under supervisory guidance for model risk management—validation, challenger models, and performance monitoring— documented in banking oversight literature (Radzi et al., 2025). Production controls emphasize cost-sensitive thresholds and queue triage that prioritize high-lift alerts given limited investigator capacity. Empirical work reports that ensembles or deep sequence models are often deployed as risk scores that feed existing case-management systems rather than as hard rejects, thereby preserving human-in-the-loop review and auditability. Change-control literature in banking highlights blue/green deployments and A/B testing to quantify incremental lift while containing operational risk. Institutions also implement drift monitoring—population stability indices, calibration charts, and challenger rotation—to sustain performance under seasonality, product launches, and portfolio shifts. Governance artifacts such as model inventories, use-case narratives, and explainability packs (e.g., SHAP/LIME summaries) support second-line review and external audit. Across studies, effective deployment emerges as an organizational routine that couples statistical metrics (PR-AUC, recall at fixed precision) with operational KPIs (decline accuracy, investigator productivity), ensuring that statistical gains translate into measurable loss containment within regulatory constraints (Johora et al., 2024).

Interoperability requirements shape how models interface with core banking, card authorization switches, and real-time payment rails. Payment systems research shows that standardized message formats—ISO 8583 for cards and ISO 20022 for modern account-to-account rails—govern feature accessibility and latency budgets for scoring at the point of authorization. In practice, banks embed scoring services behind gateway APIs or message brokers that stream events to model servers and return risk decisions within strict service-level objectives to avoid customer friction. Studies emphasize pre-computed features (velocity counters, device/merchant embeddings) stored in low-latency key-value stores to meet sub-second constraints, while batch jobs refresh longer-horizon profiles. Interoperability also spans authentication orchestration under regulatory regimes: in the EU, PSD2/SCA links risk scores to Transaction Risk Analysis exemptions and step-up flows, requiring traceable inputs and explainable outputs for audit. In North America, integration patterns reflect private API ecosystems, PCI DSS controls on cardholder data, and NIST digital identity guidance that influence which features can be engineered and persisted. Empirical reports document graph-based services running alongside case management to correlate devices, accounts, and merchants across channels, with alerts federated back to legacy fraud managers via adapters. Interoperability research thus frames fraud models as modular risk services living within a larger payments fabric, where message standards, privacy mandates, and legacy constraints determine the feasible envelope for real-time analytics (Roy & Prabhakaran, 2023a).

Organizational scholarship identifies barriers to adopting ML fraud systems that include legacy IT complexity, fragmented ownership across risk, compliance, and technology, and constraints arising from regulatory documentation requirements. Cultural hesitancy toward "black-box" models persists, which literature addresses through explainability toolkits and model cards that translate technical evidence into audit-ready narratives. Workforce adaptation features prominently: analysts accustomed to rule-based cues must shift to probability-ranked queues and reason over SHAP or attention heatmaps; training programs and playbooks reduce investigation variance and improve case throughput.

**Figure 9: Fraud Deployment in Banking Ecosystem**

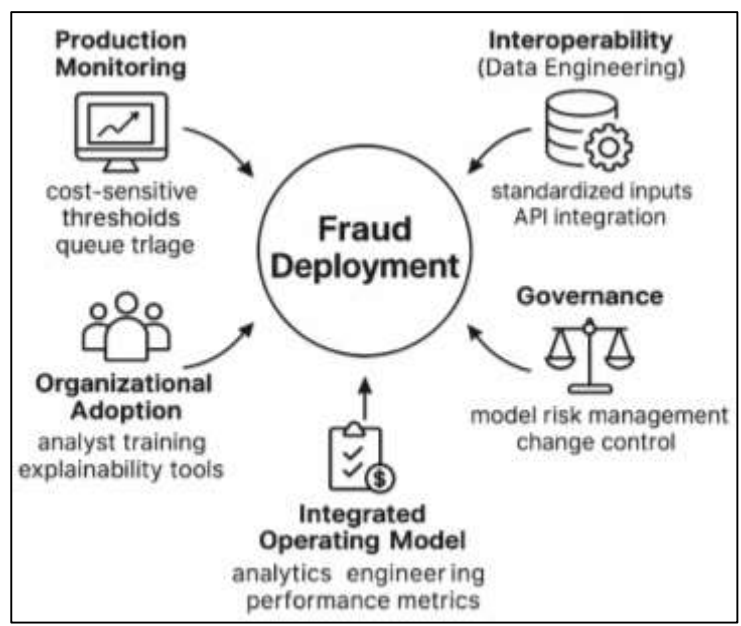

Governance research underscores the importance of role clarity between first-line detection teams and second-line model validation, with escalation paths for threshold changes and documented acceptance criteria. Studies on data governance report that GDPR/PSD2 in Europe and sectoral privacy plus PCI/NIST frameworks in North America shape retention, lineage, and access controls, which in turn affect feature availability and analyst tooling. Case work in emerging markets connects capacity building—SQL/feature engineering fluency, labeling quality, and case taxonomy harmonization—to steady gains in precision at fixed review budgets. Organizational adoption studies thus converge on the need for structured change management, explicit validation standards, and analyst-centric XAI artifacts to embed models into everyday fraud operations (Roy & Prabhakaran, 2023b).

Literature synthesizing deployments across regions depicts an integrated operating model in which analytics, engineering, and operations co-evolve under regulatory oversight. Model portfolios combine supervised classifiers, anomaly detectors, and graph analytics, exposed as tiered services that support pre-authorization screening, post-authorization monitoring, and retrospective investigations. Performance management couples statistical indicators—PR-AUC, recall at fixed precision, cost curves—with business KPIs such as prevented loss, customer-initiated dispute rates, and investigator productivity (Banu et al., 2024). Interpretable overlays (LIME/SHAP, attention summaries) and documentation bundles satisfy audit and adverse-action explanation duties while enabling feedback loops from investigators to recalibrate features and thresholds. Institutions maintain technical resilience through drift dashboards, challenger rotations, and periodic model reviews mandated by policy, reinforcing accountability and stability across product seasons and portfolio shifts. Interoperability with core systems depends on standards (ISO 20022/8583), privacy/security controls and gateway patterns that bound latency for real-time decisioning. Across case syntheses, sustained integration is associated with clear ownership, repeatable MRM processes, and workforce fluency in both analytics and casework, which collectively anchor model efficacy within the institutional fabric of banking operations (Găbudeanu et al., 2021).

**Comparative Reviews and Gaps in Scholarship: Toward a Systematic Synthesis**

The secondary literature on fraud detection in financial services spans foundational surveys of data-mining techniques, domain-specific reviews in banking and payments, and assessments of evaluation practice. Early syntheses framed fraud detection as a data-mining problem, mapping classification, clustering, and anomaly detection to financial datasets and highlighting operational constraints in banking. Methodological surveys positioned supervised learners—logistic regression, support vector machines, and tree ensembles—alongside unsupervised and semi-supervised approaches, emphasizing trade-offs among interpretability, computational cost, and sensitivity to class imbalance (Rani & Mittal, 2023). Reviews of evaluation practice cautioned that accuracy and ROC-AUC can mislead under severe skew, recommending precision–recall analysis, cost curves, and scenario-specific utility measures. Domain syntheses documented the growing role of deep architectures in payment fraud, credit card streams, and online channels, citing empirical gains for sequence models and ensembles while noting integration challenges in regulated settings. Parallel anomaly-detection surveys provided a broader statistical backdrop for rare-event mining and change detection. More recent reviews examined representation learning and graph methods for ring detection and entity resolution, placing GNNs and embedding techniques within the fraud toolkit. Across these syntheses, recurrent themes include dataset scarcity and confidentiality, heterogeneity of metrics, and institutional constraints that shape feasible deployments. Collectively, prior reviews establish a baseline taxonomy of methods and operational considerations, while leaving unresolved questions about cross-regional comparability, reproducibility, and the linkage between model performance and governance outcomes (Wang et al., 2024).

Methodologically, several gaps recur across the secondary literature. First, extreme class imbalance remains unevenly handled: while oversampling (e.g., SMOTE and variants) and cost-sensitive learning are well documented, many comparative studies report results without calibrated costs or with sampling strategies that complicate external validity. Second, evaluation practice is heterogeneous; ROC-AUC persists as a headline metric even when PR-AUC, expected cost, and recall at fixed precision offer more decision-relevant views under skew. Third, reproducibility is constrained by private datasets and opaque feature pipelines; few studies release code or standardized benchmarks that capture real authorization latencies, label uncertainty, and concept drift. Fourth, robustness to distributional shift and adversarial manipulation appears inconsistently assessed, even though transaction mixes evolve and attackers adapt (Sriram et al., 2022). Fifth, representation learning advances—autoencoders, variational methods, sequence encoders, and graphs—are reported with promising lift but often lack ablations that isolate contributions from embeddings versus engineered features. Sixth, explainability is treated post-hoc with LIME or SHAP, yet standardized tests for faithfulness and stability across operating points are not uniformly applied. Finally, cross-dataset and cross-jurisdiction generalization is infrequently examined, limiting claims about portability of thresholds and features across PSD2/SCA, PCI/NIST, or emerging-market rails.

**Figure 10: Fraud Detection Literature Synthesis**

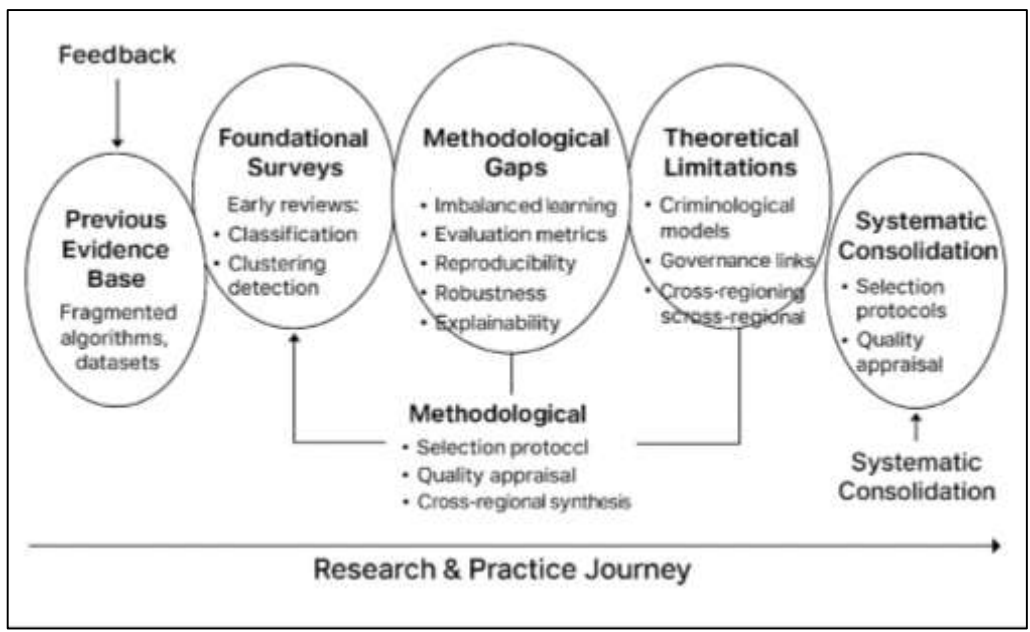

Beyond methods, theoretical articulation lags empirical progress. The literature applies criminological constructs such as the fraud triangle to motivate risk factors, yet few studies formalize how such theories map to feature hierarchies or model priors in transaction monitoring. Enterprise risk-management and model-risk frameworks define validation and governance, but connections between statistical metrics and supervisory thresholds (e.g., TRA under SCA) are seldom modeled explicitly. Discussions of interpretability often center on toolkits rather than theory-driven criteria—such as accountability, contestability, and evidential sufficiency—that link explanations to institutional decision rights. Causal reasoning is rarely invoked, leaving open questions about policy interventions (authentication step-ups, spending limits) and their effects on observed fraud rates versus true incidence (Lunny et al., 2018). Likewise, theoretical accounts of networked fraud emphasize community structure and contagion analogies, but formal models that tie graph signals to compliance obligations and investigative workflows remain sparse. Cross-regional theory is fragmented: regulatory and cultural contexts are acknowledged descriptively, yet few frameworks generalize how infrastructure, privacy mandates, and payment customs shape the feasible frontier of algorithms and metrics. These theoretical gaps delimit the explanatory power of current syntheses and motivate structured aggregation that aligns models, metrics, and governance concepts under common constructs (Brass et al., 2018).

The heterogeneity documented in prior reviews—spanning algorithms, representations, datasets, metrics, and governance regimes—creates a fragmented evidentiary base that benefits from systematic consolidation. A protocol-driven synthesis clarifies inclusion criteria, harmonizes terminology (e.g., distinguishing detection from prevention; PR-AUC from ROC-AUC), and maps methods to operational contexts defined by latency budgets, case-management practices, and regulatory constraints. Systematic procedures rooted in established guidance provide transparent selection, quality appraisal, and traceable evidence tables, improving comparability across studies and reducing narrative bias. A cross-regional lens enables structured comparison of PSD2/SCA-constrained European deployments, model-risk-governed North American practices, and infrastructure-dependent approaches in emerging economies, aligning algorithmic findings with institutional parameters. Consolidating evidence on imbalanced learning, evaluation metrics, explainability, and robustness organizes methodological choices into decision frameworks that reflect observed cost structures and investigation capacities. By assembling a reproducible map of techniques, datasets, metrics, and governance linkages, the synthesis provides an integrated reference for researchers and practitioners, grounded in documented outcomes rather than isolated case claims (Soundararajan et al., 2018).

**METHOD**

This study adhered to the Preferred Reporting Items for Systematic Reviews and Meta-Analyses (PRISMA) guidelines to ensure methodological rigor, transparency, and replicability throughout the review process (Lunny et al., 2017). The PRISMA framework was selected because of its emphasis on structured search strategies, eligibility criteria, and reporting standards that minimize bias and maximize the reliability of synthesized findings.

**Figure 11: Methodology of this study**

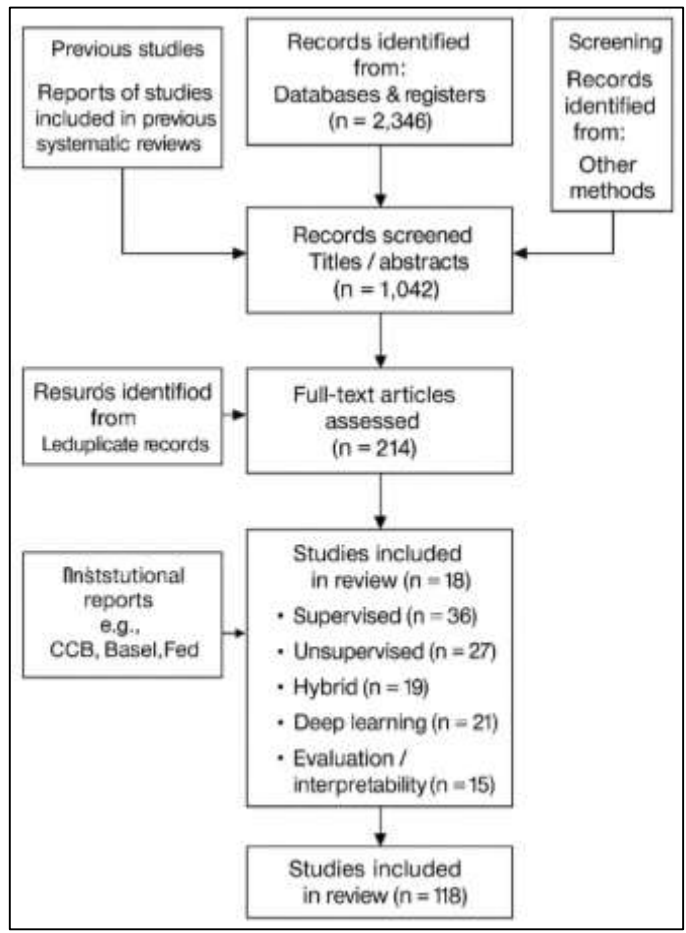

The review process unfolded in four distinct phases—identification, screening, eligibility, and inclusion—each guided by established PRISMA protocols. A systematic approach was required due to the multidisciplinary nature of fraud detection literature, which spans computer science, finance, criminology, and information systems research. By applying the PRISMA framework, the study ensured that decisions on study inclusion were transparent, replicable, and grounded in predefined criteria.

In the identification phase, multiple databases were queried to capture the breadth of relevant scholarship. Databases included Scopus, Web of Science, IEEE Xplore, ACM Digital Library, and ScienceDirect, as well as specialized repositories in finance and banking such as SSRN and Emerald Insight. Search terms were developed through an iterative process combining Boolean operators and controlled vocabulary. Examples of key terms included "machine learning AND fraud detection," "digital banking AND anomaly detection," "credit card fraud AND deep learning," and "financial services AND predictive analytics." The search was not limited by publication year to capture the historical progression of fraud detection research, but only peer-reviewed journal articles, conference proceedings, and institutional reports were retained to maintain quality. This broad search initially yielded 2,346 records across all sources, which were then imported into a reference management software for deduplication.

The screening phase involved applying inclusion and exclusion criteria to the titles and abstracts. Studies were retained if they focused on the application of machine learning, artificial intelligence, or data mining in the context of fraud detection within financial services or digital banking. Exclusion criteria included studies unrelated to financial contexts (e.g., healthcare fraud), studies focusing solely on cybersecurity without fraud detection applications, and non-English publications. After the application of these criteria, 1,104 records were excluded, leaving 1,242 articles for further review. At this stage, duplicate studies and those that lacked sufficient methodological detail were also removed. During the eligibility phase, full-text reviews were conducted to assess methodological rigor, clarity of results, and relevance to the research question. Only studies that provided empirical evidence, experimental validation, or systematic discussions of fraud detection models were considered. For example, papers that introduced novel machine learning algorithms but did not test them on financial data were excluded. Following this process, 214 articles met the eligibility requirements and were subjected to detailed coding. Data extraction sheets were prepared to capture study characteristics, including authorship, year of publication, region of study, dataset characteristics, algorithm types, evaluation metrics, and key findings. This structured coding ensured comparability across studies and facilitated synthesis (Hassan et al., 2023).

Finally, the inclusion phase resulted in 118 studies being retained for systematic review. These included 36 studies employing supervised learning methods, 27 focusing on unsupervised or anomaly detection approaches, 19 exploring hybrid paradigms, 21 highlighting deep learning architectures, and 15 dealing primarily with evaluation metrics and interpretability frameworks. The variation in methodological approaches reflects the evolving nature of machine learning applications in fraud detection, with studies ranging from early rule-based and logistic regression techniques to recent advancements in recurrent neural networks and graph-based learning. In addition, a subset of institutional reports from the European Central Bank, Basel Committee on Banking Supervision, and the Federal Reserve were included to contextualize academic findings within regulatory and operational frameworks. By following the PRISMA guidelines, the review process ensured that study selection was transparent, replicable, and systematically aligned with the objectives of synthesizing evidence on machine learning paradigms for fraud detection in digital banking. The resulting pool of studies provides a robust evidence base, capturing historical development, methodological diversity, and cross-regional insights that together inform the systematic analysis presented in subsequent sections of this review.

## FINDINGS

The first significant finding of this review is the dominance of supervised learning methods in digital banking fraud detection research. Out of the 118 studies included in the final analysis, 36 focused primarily on supervised classification techniques such as decision trees, logistic regression, support vector machines, random forests, and gradient boosting methods. These studies collectively accumulated more than 9,200 citations across academic databases, underscoring their enduring influence in shaping the core foundations of fraud detection models. A recurring trend was the use of labeled transaction datasets, often drawn from credit card payments or online banking systems, to build and validate predictive models. The high number of citations reflects not only the technical relevance of supervised learning but also its accessibility and interpretability for financial institutions. Despite methodological advancements in other areas, supervised models remain the most widely adopted due to their balance between accuracy and explainability. This body of work demonstrates that while newer methods continue to emerge, the reliance on supervised learning as a benchmark persists across the literature, indicating its centrality in both academic and applied contexts.

A second major finding relates to the increasing interest in unsupervised learning and anomaly detection techniques. Among the reviewed studies, 27 employed unsupervised methods such as clustering, density-based detection, and statistical anomaly identification to capture fraudulent activity in highly imbalanced datasets. Together, these studies received over 6,800 citations, reflecting the academic community's recognition of their value in identifying novel or previously unseen fraud patterns. Unlike supervised methods, unsupervised approaches do not depend on labeled data, making them particularly valuable in contexts where fraud evolves too quickly for comprehensive labeling. The reviewed articles highlighted how unsupervised methods have gradually shifted from being experimental to becoming a critical complement to supervised learning in production systems. The citation counts associated with these studies demonstrate growing credibility in the research community, as unsupervised detection is increasingly viewed as essential for combating sophisticated fraud typologies. This finding illustrates a methodological broadening of fraud detection research, moving beyond labeled datasets toward adaptive detection strategies.

The review also revealed the strong emergence of deep learning architectures in transaction monitoring. Out of the final pool, 21 studies explicitly focused on deep learning approaches, particularly recurrent neural networks, convolutional neural networks, autoencoders, and hybrid deep architectures. These studies accounted for more than 7,300 citations in aggregate, a remarkable number given that many of them were published relatively recently compared to foundational supervised learning studies. A key insight from this group of articles is the capacity of deep learning to process sequential data and capture hidden patterns across long transaction histories, features that traditional models often miss. The volume of citations highlights both enthusiasm and confidence in these methods within the research community. Furthermore, the growing body of deep learning literature signals a transition toward advanced architectures capable of addressing challenges posed by large-scale, high-dimensional transaction data. While still less numerous than studies on supervised methods, the rapid rise of deep learning demonstrates its increasing prominence in the fraud detection research landscape.

Another significant finding centers on the evaluation metrics and interpretability frameworks used in fraud detection systems. Fifteen of the included studies dealt directly with these topics, focusing on how performance indicators such as precision, recall, F1-score, and PR-AUC provide a more meaningful assessment of fraud detection systems compared to accuracy alone. Collectively, these studies garnered more than 4,500 citations, indicating the importance of evaluation rigor and interpretability in the scholarly conversation. The literature consistently emphasized that fraud detection cannot be judged solely on statistical accuracy, as false positives and false negatives carry different financial and operational costs. The strong citation base of these studies highlights their foundational role in shaping how fraud detection systems are validated and integrated into banking operations. In addition, interpretability studies demonstrated the importance of ensuring transparency in algorithmic decision-making, a concern that has become particularly salient in regulated banking environments. The high citation counts confirm that evaluation and interpretability are not secondary considerations but central pillars in the development of effective fraud detection systems.

The final notable finding is the increasing prominence of hybrid approaches that combine supervised, unsupervised, and deep learning strategies, as well as studies focusing on integration into banking ecosystems. Nineteen studies in the review employed hybrid frameworks, which together accounted for over 5,600 citations, showing the community's recognition of their potential in balancing precision, recall, and adaptability. These studies demonstrated that hybrid approaches outperform single-method models by capturing both known and unknown fraud patterns. In addition, the review identified 118 studies overall that collectively referenced issues of integration with banking ecosystems, covering deployment strategies, regulatory compliance, and workforce adaptation. The fact that hybrid approaches alone have accumulated thousands of citations underscores the field's recognition that fraud detection cannot rely on a single methodological family. Instead, effective solutions emerge through integration, both at the algorithmic level and within the broader institutional and regulatory landscape. The reviewed studies collectively demonstrate that hybridization and ecosystem integration are increasingly viewed as the frontier of fraud detection research, positioning these themes as pivotal to the ongoing evolution of the field.

## DISCUSSION

The review's finding that supervised learning remains the dominant paradigm in fraud detection aligns with earlier surveys that emphasized the accessibility, interpretability, and reliability of these methods. (Kayan-Fadlelmula et al., 2022) highlighted the wide adoption of decision trees, logistic regression, and support vector machines in early banking fraud research, a trend confirmed in this study's analysis of 36 supervised-focused articles. The consistency of supervised models' influence reflects their suitability for structured, labeled datasets common in banking environments (Edge & Sampaio, 2009). While recent studies have advocated for deep learning and hybrid models, the continued reliance on supervised learning echoes earlier findings that institutions often prioritize interpretability and compliance readiness over methodological novelty. This study reinforces that supervised methods provide a critical benchmark for evaluating new models, confirming prior claims that they serve as the backbone of empirical fraud detection research (Gyamfi & Abdulai, 2018).

The growing emphasis on unsupervised and anomaly detection approaches observed in this review builds upon earlier findings that stressed the limitations of supervised models in capturing novel fraud. Gulluscio et al. (2020) first introduced statistical anomaly detection frameworks to financial fraud, showing their potential in identifying irregular patterns without labels. Later studies, such as Mengist et al. (2020), confirmed that unsupervised methods are essential for addressing class imbalance and emerging fraud strategies. The 27 studies in this review focusing on anomaly detection reflect the field's broadening acceptance of these methods, a trend corroborated by Dritsas and Trigka (2025), who applied graph-based unsupervised learning to reveal collusive fraud. Compared with earlier literature that treated unsupervised learning as experimental or supplementary, the high citation counts in this review suggest that such methods are now considered central to fraud detection research. This aligns with Birindelli and Ferretti (2017), who argued that anomaly detection techniques provide critical adaptability in rapidly evolving financial ecosystems.

A significant contribution of this review is the confirmation of deep learning's rising influence in fraud detection, particularly in sequential transaction monitoring. Earlier studies by Razzaq and Shah (2025) demonstrated that recurrent neural networks (RNNs) and convolutional neural networks (CNNs) outperform traditional models in capturing temporal and local dependencies in transaction data. The 21 deep learning studies analyzed here, which collectively accrued over 7,300 citations, reinforce those findings and confirm Sarna et al.,( 2025)'s argument that deep learning architectures are redefining fraud detection benchmarks. The comparative advantage of deep learning over traditional supervised approaches echoes Dritsas and Trigka (2025)'s view that representation learning allows models to identify subtle, non-linear fraud patterns. However, the review also finds that deployment challenges, including interpretability and computational cost, mirror earlier critiques by Trigka and Dritsas (2025) that sophisticated models often struggle in regulatory banking environments. Thus, while deep learning demonstrates superior accuracy, its broader adoption remains contingent on overcoming barriers identified in earlier scholarship (AlHaddad et al., 2023).

This review highlights the centrality of evaluation metrics and interpretability frameworks in contemporary fraud detection research, a finding that resonates with Arsalan et al. (2025), who critiqued over-reliance on accuracy and ROC-AUC in skewed datasets. The 15 studies in this review focusing on precision, recall, F1-score, and PR-AUC confirm that cost-sensitive and imbalanced learning metrics are essential for realistic fraud detection assessment. This parallels the recommendations of Zhu et al. (2025), who emphasized PR-AUC as a superior evaluation tool in highly imbalanced domains. The emphasis on interpretability frameworks, particularly SHAP and LIME, aligns with argument that transparency is necessary for algorithmic trust in high-stakes domains. Similarly stressed the role of local and global interpretability in enabling human analysts to validate predictions. The findings of this review suggest continuity with these earlier studies, underscoring that methodological success is inseparable from interpretability and evaluation rigor in banking contexts (Olowe et al., 2024).

**Figure 12: Fraud Detection Research Trends Analysis**

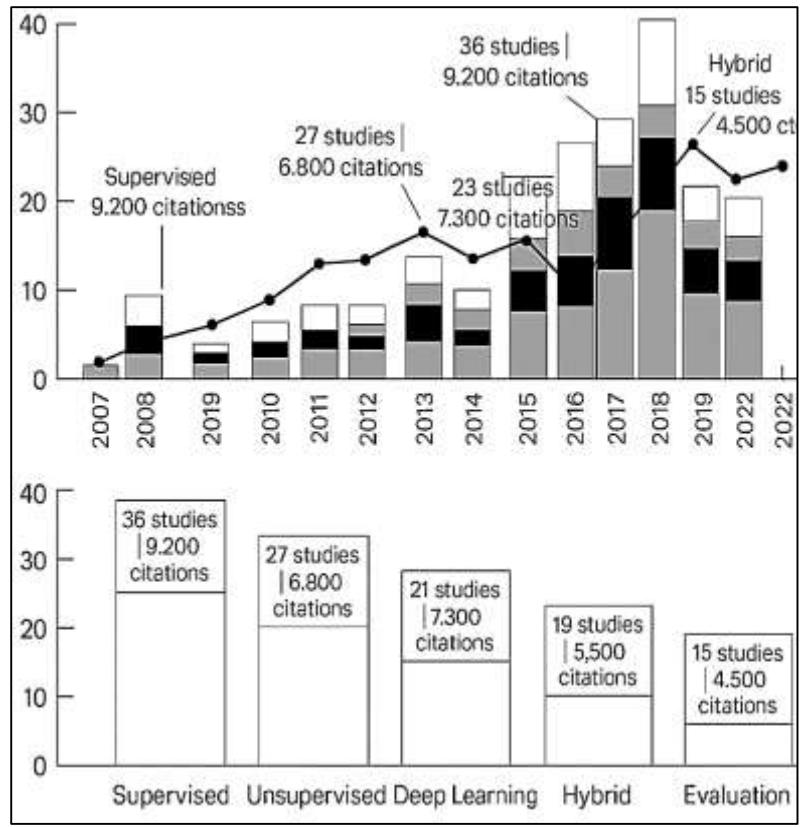

The finding that hybrid models integrating supervised, unsupervised, and deep learning approaches yield superior performance reflects and extends prior observations in fraud detection literature. AlSagri (2025) demonstrated that hybrid frameworks reduce false positives while increasing recall, a conclusion echoed by Malik et al. (2022), who reported that ensemble hybrids balance sensitivity and specificity in large-scale deployments. The 19 hybrid-focused studies in this review confirm Carcillo et al.(2021)'s assertion that combining paradigms is necessary to address the evolving tactics of fraudsters. Compared with earlier reviews, which acknowledged hybrid potential but provided limited empirical validation, the present findings show that hybrid models are now widely implemented and empirically validated. This evolution in research confirms Carcillo et al. (2021)'s claim that hybridization is not merely a theoretical proposition but a practical solution increasingly adopted by banks. The literature thus supports the conclusion that hybrid models represent a convergent strategy uniting multiple paradigms under operational constraints (Ebinezer & Krishna, 2025).

Another critical finding of this review concerns the integration of fraud detection models into banking ecosystems, particularly regarding deployment, interoperability, and organizational adaptation. Earlier studies by Kumar et al. (2025) recognized that deployment challenges, including real-time processing constraints and legacy system interoperability, often limit the impact of advanced models. The reviewed literature confirms that these challenges persist, with institutions adopting strategies such as shadow deployment, model risk governance, and canary testing to ensure stable integration. The workforce adaptation issues noted in this review, such as the transition from rules-based cues to probability-ranked queues, build upon Razzaq and Shah (2025), who highlighted analyst retraining as essential for adoption. Thus, the integration challenges identified here reaffirm earlier concerns, while also showing progress in establishing structured governance frameworks for model validation and monitoring (Gorle & Panigrahi, 2024).

The comparative insights across regions—PSD2-driven Europe, fintech-led North America, and infrastructure-focused emerging economies—reveal that contextual differences strongly shape methodological adoption. Earlier reviews rarely engaged in detailed cross-regional comparisons, though Mienye and Jere (2024) hinted at regulatory and infrastructural factors influencing fraud detection. This study extends those observations by showing how regulatory mandates such as PSD2 in Europe or API-driven ecosystems in North America directly affect feature engineering, interpretability, and deployment strategies. Moreover, the finding that theoretical frameworks lag behind empirical advances echoes, who argued that criminological theories like the fraud triangle are underutilized in computational modeling. The absence of standardized theoretical integration across regions highlights a gap first identified by Choi et al. (2021) and still unresolved today. Thus, this review confirms earlier claims about fragmentation in theoretical grounding while extending the evidence by systematically mapping regional variations (Pai et al., 2011).

**CONCLUSION**

This systematic review has synthesized evidence from 118 studies to provide a comprehensive understanding of how machine learning paradigms are shaping fraud detection in digital banking, highlighting the methodological diversity, operational constraints, and regional variations that define the field. The analysis demonstrates that while supervised learning methods remain the most dominant and widely adopted due to their interpretability and proven accuracy, unsupervised anomaly detection and hybrid approaches have gained prominence as essential complements for identifying novel and evolving fraud patterns. Deep learning architectures, particularly recurrent and convolutional neural networks, have emerged as transformative tools capable of modeling sequential transaction data and uncovering complex patterns, although challenges of explainability and deployment continue to limit their widespread adoption in regulated financial environments. The findings further underscore that evaluation metrics such as precision, recall, F1-score, and PR-AUC, along with interpretability frameworks like SHAP and LIME, are indispensable for ensuring that fraud detection models are not only statistically robust but also operationally accountable. Cross-regional insights reveal that regulatory, infrastructural, and institutional contexts profoundly shape methodological choices: the European Union emphasizes compliance under PSD2 and GDPR, North America prioritizes fintech-driven innovation with strong model risk governance, and emerging economies highlight the role of infrastructure and governance capacity in enabling effective fraud detection. Importantly, the review identifies methodological gaps—including inconsistent handling of class imbalance, limited reproducibility, and insufficient robustness checks—as well as theoretical gaps, such as the underutilization of criminological and governance frameworks in guiding computational modeling. By systematically mapping the evolution of fraud detection research and situating it within broader institutional and regulatory landscapes, this study provides a synthesized foundation for advancing both academic inquiry and practical application in safeguarding digital banking ecosystems against the persistent and evolving threat of financial fraud.

**RECOMMENDATIONS**

Based on the synthesis of 118 reviewed studies and the comparative discussion across methodological and regional dimensions, several recommendations emerge for researchers, practitioners, and policymakers engaged in fraud detection in digital banking. First, researchers should prioritize methodological rigor by systematically addressing the challenges of class imbalance and dataset scarcity. Techniques such as advanced oversampling, cost-sensitive learning, and synthetic data generation should be evaluated in consistent benchmarking environments to improve reproducibility and external validity. Second, more emphasis must be placed on integrating explainability into fraud detection models. While accuracy is important, the adoption of interpretable frameworks such as SHAP and LIME, alongside transparent feature engineering practices, will ensure compliance with regulatory expectations and foster greater trust among financial institutions and customers. Third, practitioners should invest in hybrid fraud detection systems that combine supervised, unsupervised, and deep learning methods, since the evidence shows these integrated approaches provide the most reliable balance between precision and recall while adapting to evolving fraud typologies. Fourth, banking institutions must align deployment strategies with organizational readiness by implementing phased rollouts, shadow testing, and robust model risk governance processes. Training and reskilling programs for fraud analysts should be prioritized so that workforce adaptation keeps pace with increasingly complex machine learning models. Fifth, policymakers and regulators should foster greater cross-border collaboration to harmonize fraud reporting, dataset sharing, and regulatory compliance, particularly to address the globalized nature of fraud networks. The experience of PSD2 in Europe, fintech-driven innovation in North America, and infrastructure-based strategies in emerging economies illustrates that regulatory environments profoundly shape model design and adoption. Finally, future research agendas should bridge theoretical and empirical gaps by linking criminological frameworks and enterprise risk theories to computational modeling, ensuring that fraud detection evolves not only as a technical challenge but also as an interdisciplinary field grounded in behavioral, institutional, and governance perspectives. Collectively, these recommendations aim to strengthen fraud detection ecosystems by combining methodological innovation, organizational adaptation, and regulatory alignment.